\documentclass[letterpaper, 10 pt, conference]{ieeeconf}  

\IEEEoverridecommandlockouts                              

\usepackage{times}
\usepackage{balance}
\usepackage{cite}
\usepackage{amsmath,amssymb,amsfonts}
\usepackage{algorithm}
\usepackage{algpseudocode}
\usepackage{graphicx}
\usepackage{textcomp}
\usepackage{mathtools}
\usepackage{enumerate}
\usepackage{bm}
\usepackage{makecell,multirow,diagbox}
\usepackage{xcolor}
\usepackage{subfigure}
\usepackage{threeparttable}
\usepackage{booktabs}
\usepackage[symbol]{footmisc}
\usepackage{hyperref}

\makeatletter
\algrenewcommand\ALG@beginalgorithmic{\footnotesize} 
\makeatother

\newif\ifcomments
\commentstrue
\ifcomments
\newcommand\todos[1]{\textcolor{red}{#1}}

\newcommand{\etc}[1]{~\textit{etc.}}
\newcommand{\etal}[1]{~\textit{et~al.}}

\newcommand\rso{\bgroup\markoverwith{\textcolor{red}{\rule[0.5ex]{2pt}{0.4pt}}}\ULon}

\usepackage[normalem]{ulem}
\else
\newcommand\todos[1]{}
\newcommand{\etc}[1]{}
\newcommand{\etal}[1]{}

\fi

\newcommand{\tr}{^{\!\top}}

\newcommand{\argmin}{\operatornamewithlimits{argmin}}

\newcommand{\se}{\mathfrak{se}(3)}
\newcommand{\SE}{SE(3) }

\title{\LARGE \bf
Robust Ego and Object 6-DoF Motion Estimation and Tracking
}

\author{Jun Zhang$^{1}$  Mina Henein$^{1}$  Robert Mahony$^{1}$  Viorela Ila$^{2}$
\thanks{\footnotesize $^{1}$Jun Zhang, Mina Henein and Robert Mahony are with the
        Australian National University (ANU), 0020 Canberra, Australia.
        {\tt \{jun.zhang2,mina.henein,robert.mahony\}@anu.edu.au}}%
\thanks{\footnotesize $^{2}$Viorela Ila is with the University of Sydney (USyd), 2006 Sydney, Australia.
        {\tt viorela.ila@sydney.edu.au}}%
}

\begin{document}

\maketitle
\thispagestyle{empty}
\pagestyle{empty}

\begin{abstract}

The problem of tracking self-motion as well as motion of objects in the scene using information from a camera is known as multi-body visual odometry and is a challenging task.
This paper proposes a robust solution to achieve accurate estimation and consistent track-ability for dynamic multi-body visual odometry.
A compact and effective framework is proposed leveraging recent advances in semantic instance-level segmentation and accurate optical flow estimation.
A novel formulation, jointly optimizing \SE motion and optical flow is introduced that improves the quality of the tracked points and the motion estimation accuracy.
The proposed approach is evaluated on the virtual KITTI Dataset and tested on the real KITTI Dataset, demonstrating its applicability to autonomous driving applications.
For the benefit of the community, we make the source code public\footnote[2]{\url{https://github.com/halajun/multimot_track}}.

\end{abstract}

\section{Introduction}
\label{sec:intro}

Visual odometry (VO) has been a popular solution for robot navigation in the past decade due to its low-cost and widely applicable properties.
Studies in the literature have illustrated that VO can provide accurate estimation of a camera trajectory in largely static environment, with relative position error ranging from $0.1\%$ to $2\%$~\cite{scaramuzza2011ra}.
However, the deployment of robotic systems in our daily lives requires systems to work in significantly more complex, dynamic environments.
Visual navigation in non-static environments becomes challenging because the dynamic parts in the scene violate the motion model of camera.
If moving parts of a scene dominate the static scene, off-the-shelf visual odometry systems either fail completely or return poor quality trajectory estimation.
Earlier solutions proposed to directly remove the dynamic information via robust estimation~\cite{tan2013ismar,alcantarilla2012icra}, however, we believe that this information is valuable if it is properly used.
In most scenarios, the dynamics corresponds to a finite number of individual objects that are rigid or piecewise rigid, and their motions can be tracked and estimated in the same way as the ego-motion.
Accurate object motion estimation and tracking becomes highly relevant in many applications, such as collision avoidance in autonomous driving and robotic systems, visual surveillance and augmented reality.

\begin{figure}[h]
 \vspace*{3mm}
 \centering
 \includegraphics[width=.98\columnwidth]{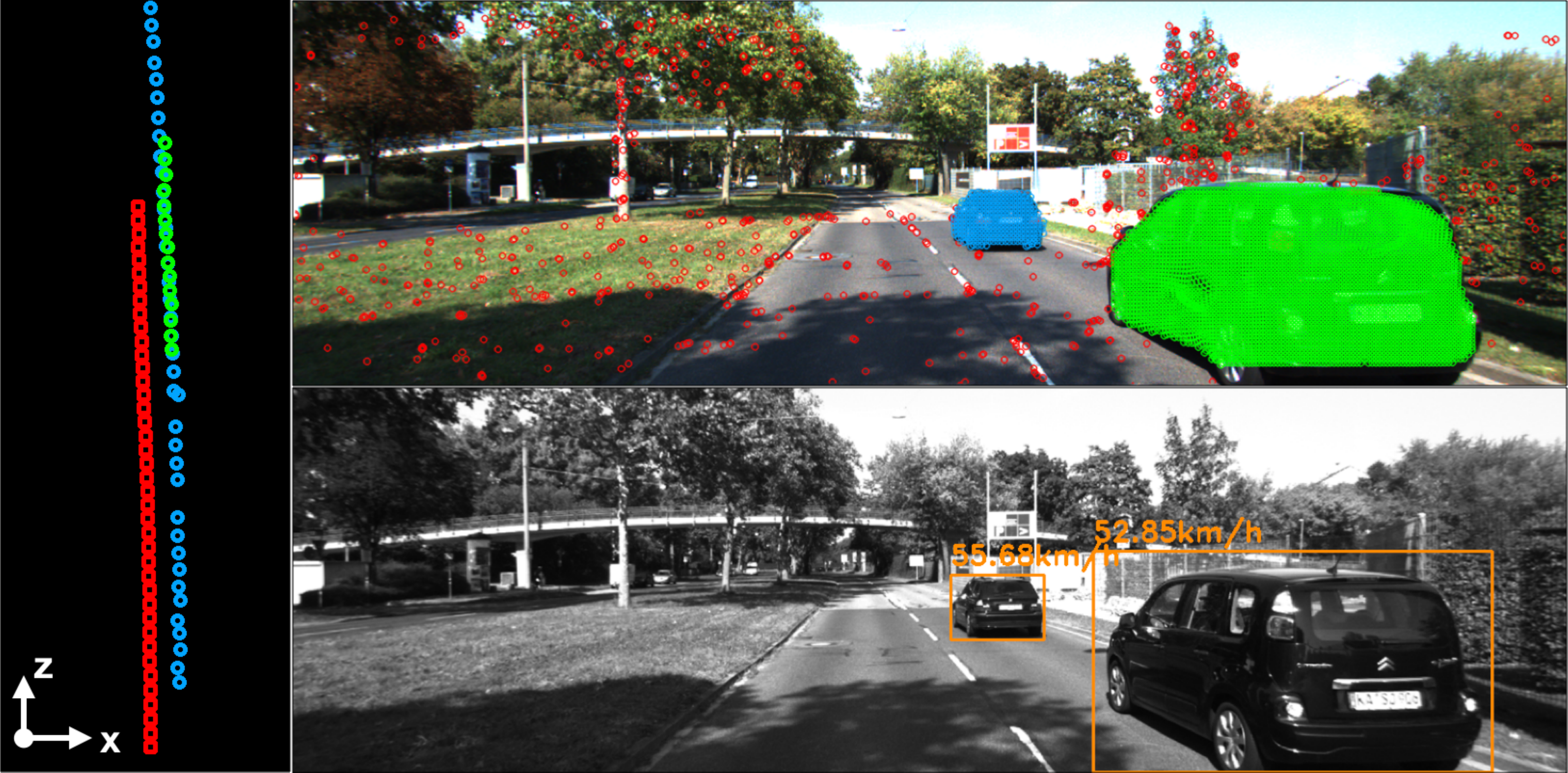}
 \caption{Results of our proposed system on KITTI sequence $03$. Camera and object trajectory (left). Detected points on background and object body (upper-right). Estimated object speed (bottom-right). }
 \label{fig:showcase}
 \vspace*{-5mm}
\end{figure}

In this paper, we propose a novel multi-body visual odometry pipeline that address the problem of tracking both ego and object motion in dynamic outdoor scenes.
The proposed pipeline leverages instance-level object segmentation algorithms~\cite{he2017iccv} to robustly separate the scene into static background and multiple dynamic objects. 
Recent advances in optical flow estimation~\cite{Ilg2017cvpr,sun2018cvpr} are exploited to maintain enough tracking points on each object to accurately estimate motion.
With this data, we propose a new technique that jointly refines the initial optical flow and estimates full $6$-DoF motion of both the camera and objects in the scene.
We construct a fully-integrated system that is able to robustly estimate and track self and object motions utilizing only visual sensors (stereo/RGB-D).
To the best of our knowledge, our work is the first to conduct an extensive evaluation of accuracy and robustness of ego and object $6$-DoF motion estimation and tracking, and demonstrates the feasibility on real-world outdoor datasets.

In the following, after Sec.~\ref{sec:related_work} on related work, we introduce the methodology of our proposed algorithm in Sec.~\ref{sec:method}, then describe the implementation of proposed pipeline in Sec.~\ref{sec:imple}.
Experimental results are documented in Sec.~\ref{sec:experi}.

\section{Related Work}
\label{sec:related_work}

Visual odometry/SLAM for dynamic environments has been actively studied in the past few years, as described in a recent survey~\cite{saputra2018csur}.
Earlier approaches detected non-static object in the scene and removed them from the estimation data.
For instance, \cite{alcantarilla2012icra} uses dense scene flow for dynamic objects detection, and obtains improved localization and mapping results by removing erroneous measurements on dynamic objects from estimation.
The authors in~\cite{tan2013ismar} propose an online keyframe update that reliably detects changed features by projecting them from keyframes to current frame for appearance and structure comparison, and discards them if necessary.

Meanwhile, researchers have started to incorporate dynamic information into camera pose estimation.
A multi-camera SLAM system is proposed in~\cite{zou2013pami}, that is able to track multiple cameras, as well as to reconstruct the 3D positions of both static background and moving foreground points.
The idea is that points on moving objects give information about relative poses between different cameras at the same time step.
Therefore, static and dynamic points are used together to decide all camera poses simultaneously.
Kundu~\cite{kundu2011iccv} proposed to detect and segment motion using efficient geometric constraints, then reconstruct the motion of dynamic objects with a bearing only tracking.
Similarly, a multi-body visual SLAM framework is introduced in~\cite{reddy2016iros}, which makes use of sparse scene flow to segment moving objects, then estimate the poses of camera as well as moving objects, respectively.
Poses are formulated as a factor graph incorporating with constraints to reach a final optimization result.

Lately, the problem of object motion estimation and tracking is receiving increased attention in the robotics and computer vision community.
Dewan~\cite{dewan2016icra} presents a model-free method for detecting and tracking moving objects in 3D LiDAR scans by a moving sensor.
The method sequentially estimates motion models using RANSAC~\cite{fischler1981acm}, then segments and tracks multiple objects based on the models by a proposed Bayesian approach.
Results of sensor/objects speed error are illustrated to prove its effectiveness.
In~\cite{judd2018iros}, the authors address the problem of simultaneous estimation of ego and third-party motions in complex dynamic scenes using cameras.
They apply multi-model fitting techniques into a visual odometry pipeline to estimate all rigid motions within a scene.
Promising results of \SE motions have been shown on multiple moving cubes dataset for indoor scenes.

\section{Methodology}
\label{sec:method}

\begin{figure*}[h]
    \centering
    \subfigure{\includegraphics[height=1.4in]{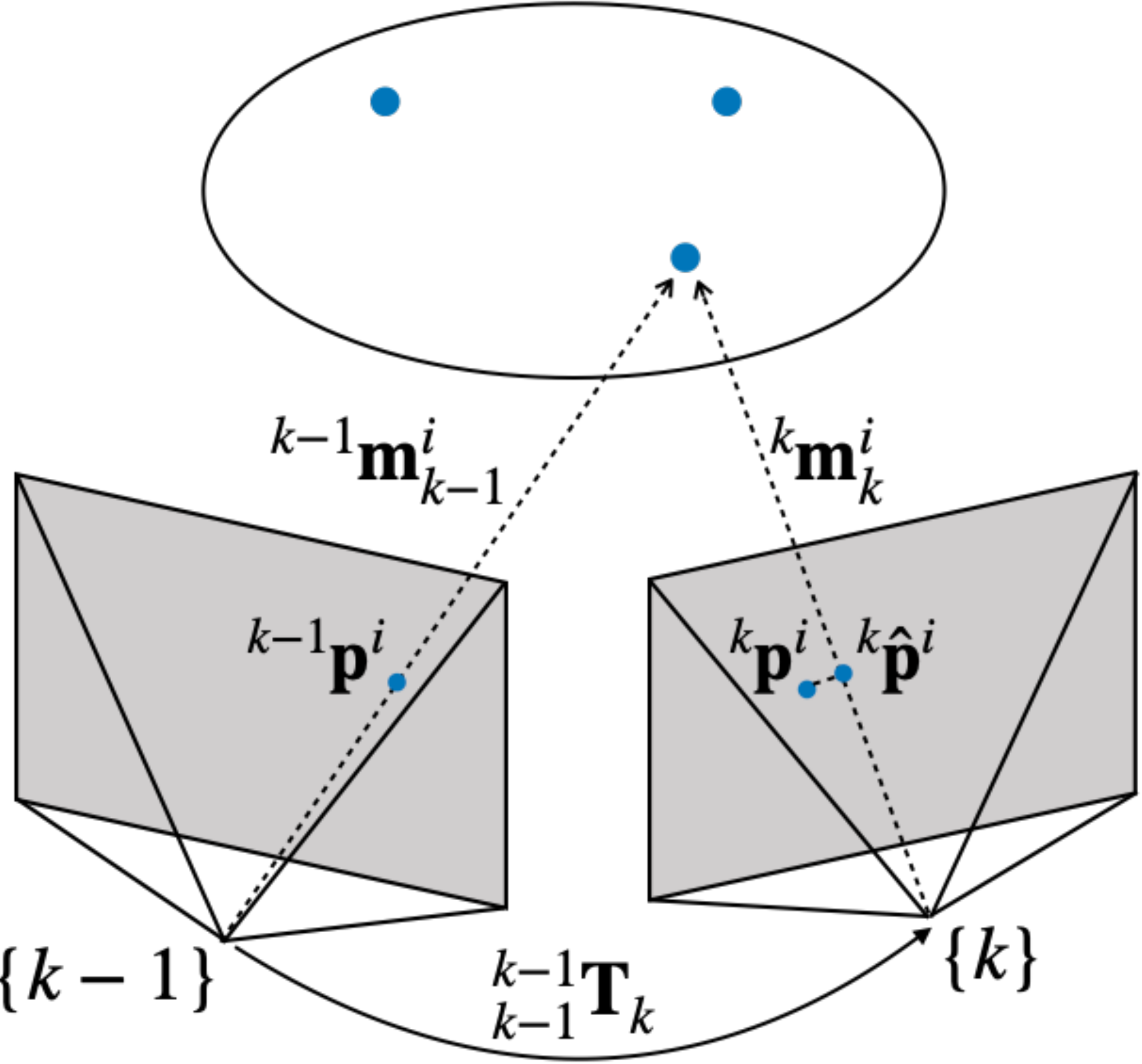}}
    \hspace{5em}
    \subfigure{\includegraphics[height=1.4in]{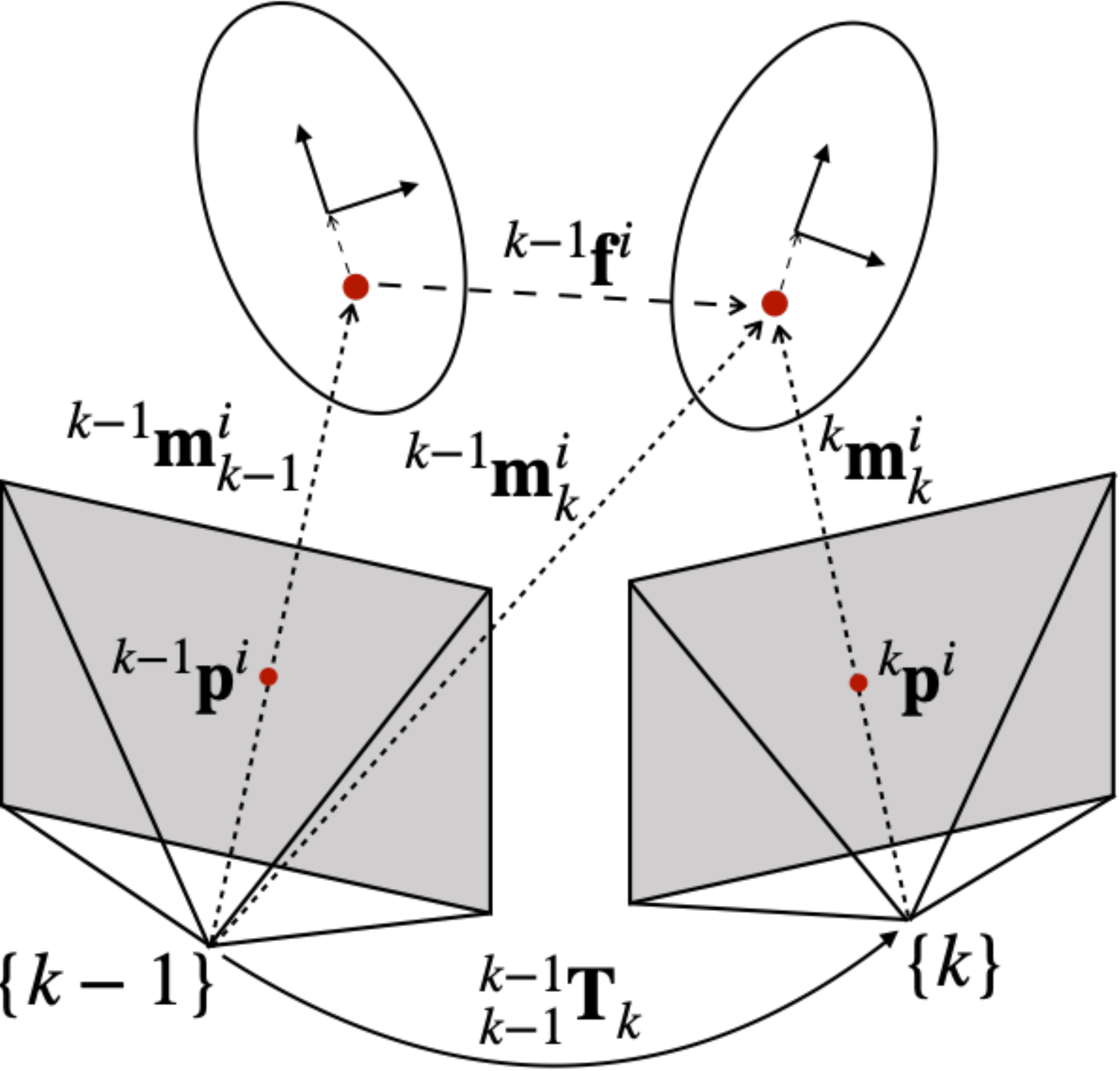}}
    \hspace{5em}
    \subfigure{\includegraphics[height=1.4in]{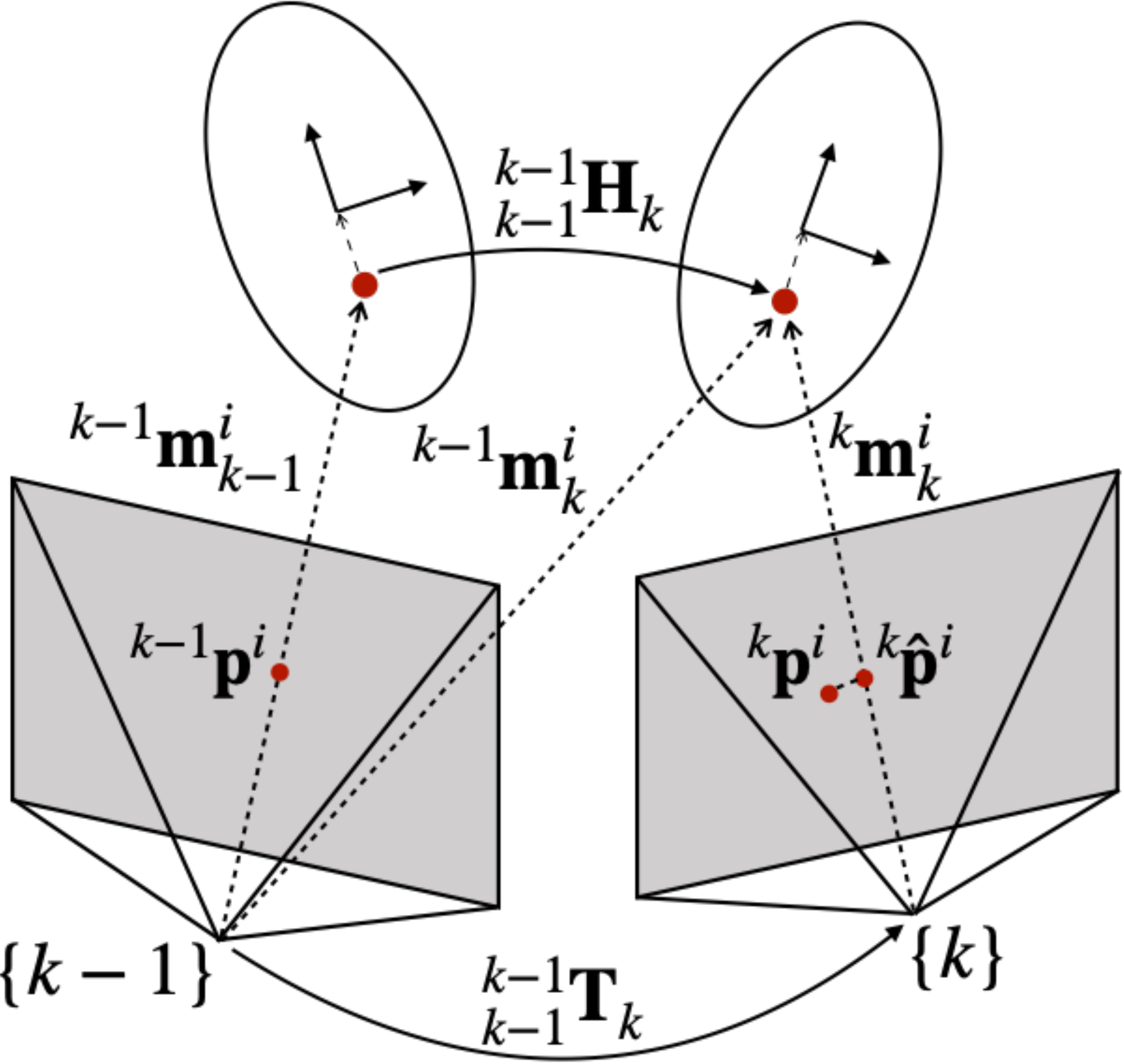}}
    \caption{Sketch maps of ego-motion obtained from static points (left), scene flow of points on moving objects (center) and rigid motion of points on moving object (right). Here blue dots represent static points, and red dots dynamic points.}
    \label{fig:mot}
    \vspace*{-4mm}
\end{figure*}

Our setup comprises a depth camera (stereo or RGB-D) moving in a dynamic environment.
Let \mbox{${}^{k}\bm{P} = \{{}^{k}\bm{p}^{i} \in \mathbb{R}^3\}$} be a set of projected points into the image frame $k$, where ${}^{k}\bm{p}^{i} = [{u}^{i},\: {v}^{i}\:, 1]\tr$ represents the point location in homogeneous coordinates.
The points are either part of the static background \mbox{ ${}^{k}\bm{P}_{s}\subseteq {}^{k}\bm{P}$} or moving object $ {}^{k}\bm{P}_{o} \subset {}^{k}\bm{P} $.

Assuming that a depth map \mbox{${}^{k}\bm{D} = \{{}^{k}d^{i} \in \mathbb{R}$\}} of frame $k$ is provided, where ${}^{k}d^{i}$ is the corresponding depth for each point ${}^{k}\bm{p}^{i} \in {}^{k}\bm{P}$, the 3D point ${}^{k}\bm{m}^{i} \in \mathbb{R}^4$ of ${}^{k}\bm{p}^{i}$ can be obtained via back-projection:
\begin{eqnarray}
\label{eq:backproj}
{}^{k}\bm{m}^{i} =
\begin{bmatrix}
    {m}_{x}^{i} \\
    {m}_{y}^{i} \\
    {m}_{z}^{i}  \\
    1
    \end{bmatrix}
= \pi^{-1}({}^{k}\bm{p}^{i})
= \begin{bmatrix}
    (u^{i}-c_u)\cdot{}^{k}d^{i}/f  \\
    (v^{i}-c_v)\cdot{}^{k}d^{i}/f  \\
    {}^{k}d^{i} \\
    1
    \end{bmatrix}
\end{eqnarray}
\noindent where $\pi^{-1}(\cdot)$ is the inverse of projection function, $f$ the focal length and $(c_{u},c_{v})$ the principal point of the cameras.

The motion of the camera between frames $k-1$ and $k$ and/or the motion of objects in the scene produce an optical flow \mbox{${}^{k}\bm{\Phi} = \{{}^{k}\bm{\phi}^{i} \in \mathbb{R}^2$\}}, where ${}^{k}\bm{\phi}^{i}$ is the corresponding optical flow for each point ${}^{k}\bm{p}^{i}$ and its correspondence ${}^{k-1}\bm{p}^{i}$ in frame $k-1$ and is given by:
\begin{eqnarray}
\label{eq:of}
{}^{k}\bar{\bm{p}}^{i} = {}^{k-1}\bar{\bm{p}}^{i} + {}^{k}\bm{\phi}^{i}\:
\end{eqnarray}
where ${}^{k}\bar{\bm{p}}^{i}$ and ${}^{k-1}\bar{\bm{p}}^{i}$ only contain the 2D point coordinates. \mbox{${}^{k}{\bm{\Phi}}$} can be obtained using off-the-shelf classic or learning-based methods.
The motions of the camera and objects in the scene are represented by pose change transformations. The following subsections will describe our new approach to estimate those.

\subsection{Camera Motion Estimation}
\label{subsec:CameraMotion}
The camera motion between frame $k-1$ (lower left index) and $k$ (lower right index) represented in body-fixed frame $k-1$ (upper left index) is denoted $^{k- 1}_{k- 1}{\bf{T}}_{k} \in \SE$. 
The image plane points, associated with static 3D points ${}^{k-1}\bm{m}_{k-1}^{i}$, observed at time $k-1$, by the projection onto the $k$ image plane can now be computed by 
\begin{eqnarray}
\label{eq:proj_sta}
{}^{k}\bm{\hat{p}}^{i} :=  \pi({}^{k}\bm{m}_{k-1}^{i}) = \pi({}^{k-1}_{k-1}\bm{T}^{-1}_{k}\:{}^{k-1}\bm{m}_{k-1}^{i}).
\end{eqnarray}
Parameterize the $\SE$ camera motion by elements \mbox{$\bm{\xi}_k^\wedge \in \se$} the Lie-algebra of $\SE$. 
That is 
\begin{eqnarray}
{}^{k-1}_{k-1}{\bf{T}}_{k} = \exp({}_{k-1}^{k-1}{\bm{\xi}}_{k}^{\wedge})  
\end{eqnarray}
where ${}_{k-1}^{k-1}{\bm{\xi}}_{k} \in \mathbb{R}^6$ and the wedge operator is the standard lift into $\se$.  
Combining \eqref{eq:of} and \eqref{eq:proj_sta}, and using the Lie-algebra parameterization of $\SE$ 
the minimizing solution of the least squares cost criteria we consider is given by 
\begin{eqnarray}
\label{eq:cost_1}
{}_{k-1}^{k-1}{\bm{\xi}}_{k}^{*} = \underset{{}_{k-1}^{k-1}{\bm{\xi}}_{k}}{\argmin} \sum_{i=1}^{n_{s}}{ \rho_{h}(|| {}^{k-1}\bar{\bm{p}}^{i} + {}^{k}\bm{\phi}^{i} - {}^{k}\hat{\bar{\bm{p}}}^{i} ||^{2}_{\Sigma_1}) }
\end{eqnarray}
for all the visible 3D-2D static point correspondences \mbox{$i ={1,...,n_{s}}$}. 
Here $\rho_{h}$ is the Huber robust cost function, and $\Sigma_1$ is covariance matrix associated to the re-projection threshold used in initialization.
The estimated camera motion is given by ${}^{k-1}_{k-1}{\bf{T}}_{k}^* = \exp({}_{k-1}^{k-1}{\bm{\xi}}_{k}^{*})$ and is found using the Levenberg-Marquardt algorithm to solve for \eqref{eq:cost_1}.

\subsection{Moving Points Motion Estimation}
\label{subsec:MotionModel}
In this section we derive the motion model of 3D points on a rigid body in motion. The motion of the rigid body in body-fixed frame is given by  \mbox{${}^{L_{k-1}}_{k-1}\bm{H}_{k}\in \SE$}. 
If the pose of the object at time $k-1$ in global reference frame  is given by ${}^{0}\bm{L}_{k-1}\in \SE$, we showed in~\cite{zhang2018acra} and~\cite{henein2020icra} that the \emph{rigid body pose transformation in global frame} is given by 
\begin{eqnarray}
_{k-1}^{0} \bm{H}_{k} = ^{0}\bm{L}_{k-1} \: _{k-1}^{L_{k-1}} \bm{H}_{k} \: ^{0}\bm{L}_{k-1}^{-1} \in \SE. 
\label{eq:Hinertial}
\end{eqnarray}
Consequently the motion of a point on a rigid body in global frame is given by $_{k-1}^{0} \bm{H}_{k}$, with the following relation:
\begin{eqnarray}
\label{eq:pochange}
{}^{0}\bm{m}^{i}_{k} = {}^{0}_{k-1}\bm{H}_{k}\:{}^{0}\bm{m}^{i}_{k-1}\:.
\end{eqnarray}
When formulating the motion estimation problem considering only two consecutive frames, the motion in the global frame in \eqref{eq:Hinertial} would be expressed
in the image frame $k-1$,  and is denoted $_{k-1}^{k-1} \bm{H}_{k}$.

As shown in Fig.~\ref{fig:mot} (right), a 3D point ${}^{k-1}\bm{m}^{i}_{k-1}$ observed on a moving object at time $k-1$, moves according to \eqref{eq:pochange} to \mbox{${}^{k-1}\bm{\hat m}^{i}_{k} = {}^{k-1}_{k-1}\bm{H}_{k}\:{}^{k-1}\bm{m}^{i}_{k-1}$}. 
The projection of the estimated 3D point onto the image frame at time $k$ is given by 
\begin{eqnarray}\nonumber
{}^{k}\hat{\bf{p}}^{i} := & \pi \left( {_{k-1}^{k-1}\bm{T}^{-1}_{k} \:  {}^{k-1}_{k-1}\bm{H}_{k}\:{}^{k-1}\bm{m}^{i}_{k-1}
}\right ) \\
 = & \pi \left( _{k-1}^{k-1}\bm{X}_{k}\:{}^{k-1}\bm{m}^{i}_{k-1} \right )\label{eq:proj_dyn}
\end{eqnarray}
\noindent where \mbox{$_{k-1}^{k-1}\bm{X}_{k} \in \SE$}. Similar to the camera motion estimation, we parameterize \mbox{${}^{k-1}_{k-1}{\bf{X}}_{k} = \exp \left ( ^{k-1}_{k-1}{\bm{\zeta}}_{k} ^{\wedge}\right)$}, with \mbox{${}^{k-1}_{k-1}{\bm{\zeta}}_{k} ^{\wedge}$} the $\se$ representation of \mbox{$^{k-1}_{k-1}{\bm{\zeta}}_{k} \in \mathbb{R}^6$}, and find the optimal solution via minimizing
\begin{eqnarray}
{}^{k-1}_{k-1}\bm{\zeta}_{k}^{*} = \underset{{}^{k-1}_{k-1}\bm{\zeta}_{k}}{\argmin} \sum_{i=1}^{n_o}{ \rho_{h}(||{}^{k-1}\bar{\bm{p}}^{i} + {}^{k}\bm{\phi}^{i} - {}^{k}\hat{\bar{\bm{p}}}^{i} ||^{2}_{\Sigma_1}) }
\label{eq:cost_2}
\end{eqnarray}
given all the $3$D-$2$D point correspondences on an object \mbox{$i = {1,...,n_{o}}$}. The motion of the object points, \mbox{${}^{k-1}_{k-1}\bm{H}_{k} = {}^{k-1}_{k-1}\bm{T}_{k}\:{}^{k-1}_{k-1}\bm{X}_{k}$} can be recovered afterwards.

\subsection{Refining the estimation of the optical flow}
\label{subsec:OpticalFlow}
Both, camera motion and object motion estimations rely on good image correspondences.
Tracking of points on moving objects can be very challenging due to occlusions, large relative motions and large camera-object distances.
In order to insure a robust tracking of points, the technique proposed in this paper aims at refining the estimation of the optical flow jointly with the motion estimation:
\begin{eqnarray}
\label{eq:cost_3}
\begin{aligned}
\{{\bm{\theta}}^{*},{}^{k}{\bm{\Phi}}^{*}\} =
& \underset{\{{\bm{\theta}},{}^{k}{\bm{\Phi}}\}}{\argmin} \sum_{i=1}^{n}\rho_{h}({ || {}^{k-1}\bar{\bm{p}}^{i} + {}^{k}\hat{\bm{\phi}}^{i} - {}^{k}\hat{\bar{\bm{p}}}^{i} ||^{2}_{{\Sigma}_{1}})} \\
& + \rho_{h}(||{}^{k}{\bm{\phi}}^{i} - {}^{k}\hat{\bm{\phi}}^{i}||^{2}_{{\Sigma}_{2}})
\end{aligned}
\end{eqnarray}
\noindent where $\{{\bm{\theta}},{}^{k}{\bm{\Phi}}\}$ can be either \mbox{$\{{}^{k-1}_{k-1}{\bm{\xi}}_{k},{}^{k}{\bm{\Phi}}_{s}\}$} for camera motion estimation, or \mbox{$\{{}^{k-1}_{k-1}{\bm{\zeta}}_{k},{}^{k}{\bm{\Phi}}_{o}\}$} for the object motion estimation, with \mbox{ ${}^{k}\bm{\Phi}_{s}\subseteq {}^{k}\bm{\Phi}$} and  \mbox{${}^{k}\bm{\Phi}_{o} \subset {}^{k}\bm{\Phi}$}. 
Here $\Sigma_{2}$ is the covariance matrix associated to initial optic-flow obtained using classic or learning-based methods. 

\begin{figure}[h]
 \centering
 \includegraphics[width=.98\columnwidth]{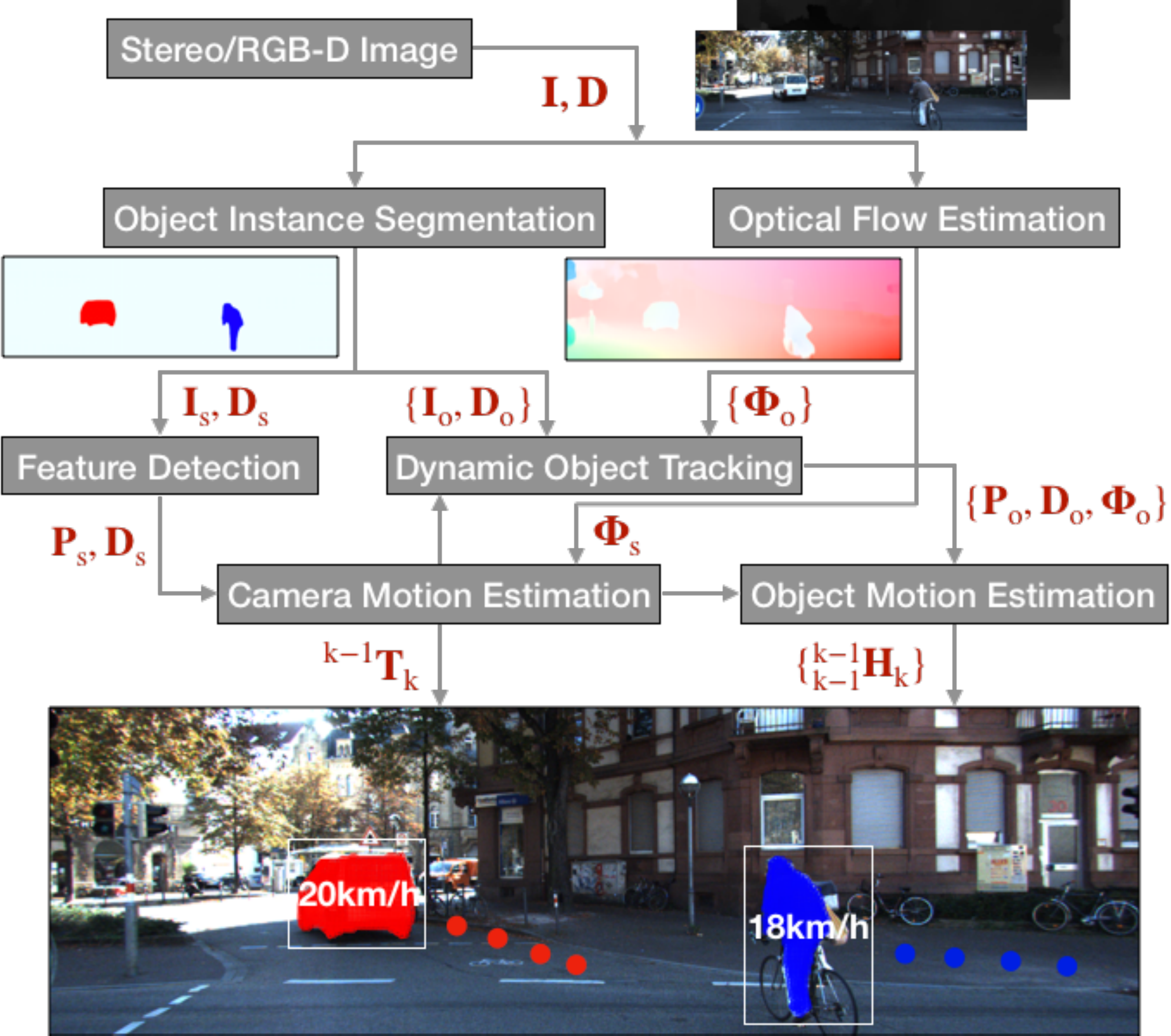}
 \caption{Overview of our multi-motion visual odometry system. Letters in red colour refer to output for each blocks. $\{\cdot\}$ denotes multiple objects.}
 \label{fig:pipeline}
 \vspace{-.4cm}
\end{figure}

\section{Implementation}
\label{sec:imple}

\noindent In this section, we propose a novel multi-motion visual odometry system that robustly estimates both camera and object motions.
Our proposed system takes stereo or RGB-D images as input.
If the input data is stereo images, we can apply the method in \cite{yamaguchi2014cvpr} to generate the depth map $\bm{D}$.
The proposed pipeline is summarised in Fig.~\ref{fig:pipeline} and contains three main parts: image preprocessing, ego-motion estimation and object motion tracking.

\subsection{Image Preprocessing}
There are two challenging aspects that this pipeline needs to fulfill.
One is to separate static background and objects, and the other is to ensure long-term tracking of dynamic objects.
For that, we leverage recent advances in computer vision technics for dense optical flow calculation and instance level semantic segmentation in order to ensure good object tracking and efficient object motion segmentation.

The dense optical flow is used to maximize the number of track points on moving objects.
Most of the moving objects, they only occupy a small portion of the image. Therefore, using sparse feature matching does not guarantee a robust feature tracking.
Our approach makes use of dense optical flow to considerably increase the number of object points.
At the same time, our method enhances the matching performance by refining optical flow jointly within the motion estimation process as presented in Section~\ref{subsec:OpticalFlow}.

Instance-level semantic segmentation is used to segment and identify potentially movable objects in the scene.
Semantic information constitutes an important prior in the process of separating static and moving object points, e.g., buildings and roads are always static, but cars can be static or dynamic.
Instance segmentation helps to further divide semantic foreground into different instance masks, which makes it easier to track each individual object.
Moreover, segmentation mask provides precise boundary of the object body that ensures robust tracking of points on objects.

The image preprocessing part of the pipeline generates the image mask, the depth and the dense flow for the static $\mathrm{{\bf{I}}_{s}}$, $\mathrm{{\bf{D}}_{s}}$ and $\mathrm{{\bf{\Phi}}_{s}}$ and dynamic $\{\mathrm{{\bf{I}}_{o}},\mathrm{{\bf{D}}_{o}},\mathrm{{\bf{\Phi}}_{o}}\}$ parts of the scene.



\subsection{Ego-motion Estimation}

To achieve fast ego-motion estimation, we construct a sparse feature set $\bm{P}_{s}$ in each frame.
Since dense optical flow is available, we use optical flow to match those sparse features across frames.
Those sparse features are only detected on regions of the image other than labeled objects.
To ensure robust estimation, a motion model generation method is applied for initialisation.
Specifically, the method generates two models and compares their inlier numbers based on re-projection error.
One model is generated by propagating the previous camera motion, while the other by computing a new motion transform using P$3$P~\cite{ke2017cvpr} algorithm with RANSAC.
The motion model that produces most inliers is then selected for initialisation.
\subsection{Object Motion Tracking}

The process of object motion tracking consists of three steps. The first step is to classify all the objects into dynamic and static objects. Then we associate the dynamic objects across the two frames. Finally, individual object motion is estimated.

\subsubsection{Classifying Dynamic Object}

Instance level object segmentation allows us to separate objects from background.
Although the algorithm is capable of estimating the motions of all the segmented objects, dynamic object identification helps reduce computational cost of the proposed system.
This is done using scene flow estimation as shown in Fig.~\ref{fig:mot} (center).
Specifically, after obtaining camera motion ${}^{k-1}_{k-1}\bm{T}_{k}$, the scene flow vector ${}^{k-1}\bm{f}^{i}$ describing the motion of a 3D point ${}^{k-1}\bm{m}_{k-1}^{i}$ between frame $k-1$ and $k$, can be calculated as~\cite{lv2018eccv}:
\begin{eqnarray}
\label{eq:sf}
{}^{k-1}\bm{f}^{i} = {}^{k-1}\bm{m}_{k-1}^{i} - ({}^{k-1}_{k-1}\bm{T}_{k}\:{}^{k}\bm{m}_{k}^{i})
\end{eqnarray}
Unlike optical flow, the scene flow can directly decide whether the scene structure is moving or not. Ideally, the magnitude of the scene flow vector should be zero for static 3D point. However, noise or error in depth and matching complicates the situation in real scenarios.

To robustly tackle this, we compute the scene flow magnitude of all the sampled points on each object, and separate them into two sets (static and dynamic) via thresholding.
An object is recognised dynamic if the proportion of ``dynamic'' points is above a certain level, otherwise static.
Table~\ref{tab:dyn_obj} demonstrates the performance of classifying dynamic and static objects using this strategy.
Overall, the proposed approach achieves good accuracy among the tested sequences.
Notice that, in sequence $20$, we have relatively high false negative cases. That is because most cars throughout sequence $20$, move slowly (nearly static) due to traffic jams.

\begin{table}[h]
  \centering
  \fontsize{7}{8}\selectfont
  \caption{Performance of dynamic/static object classification over virtual KITTI dataset.}
  \label{tab:dyn_obj}
 \begin{tabular}{lccccc}
  \toprule
  Sequence            &01        &02     &06     &18     &20     \cr
  \midrule
  Total Detection     &1383      &150    &266    &970    &2091     \cr
  Dynamic/Static      &117/1266  &73/77  &257/9  &970/0  &1494/597 \cr
  False Positive      &3         &0      &9      &0      &3        \cr
  False Negative      &6         &0      &0      &57     &292      \cr
  \bottomrule
 \end{tabular}
 \vspace{-.4cm}
\end{table}

\subsubsection{Object Tracking}

Instance-level object segmentation only provides labels frame by frame, therefore objects need to be tracked between frames and their motion models propagated over time.
To manage this, we propose to use optical flow to associate point labels in accross frames.
For that, we introduce and maintain a finite tracking label set \mbox{$\mathcal{L}\subset \mathbb{N}$} where \mbox{$l \in \mathcal{L}$} starts from $l=1$, when the first moving object appears in the scene.
The number of elements in $\mathcal{L}$ increases as more objects are being detected.
Static objects and background are labeled with \mbox{$l=0$}.

Ideally, for each detected object in frame $k$, the labels of all its points should be uniquely aligned with the labels of their correspondences in previous frame $k-1$.
However, in practice this is affected by the noise, image boundary and occlusions.
To overcome this, we assign all the points with the label that appears most in their correspondences.
For a dynamic object, if the most frequent label in the previous frame is $0$, it means that the object starts to move, appears in the scene at the boundary, or reappears from occlusion.
In this case, the object is assigned with a new tracking label.

\subsubsection{Object Motion Estimation}

As mentioned before, objects normally appear in small sizes in the scene, which makes it hard to get sufficient sparse features to track and estimate their motions robustly.
Therefore we densify the object point set ${\bm{P}}_{o}$ via sampling every $3^{rd}$ pixel within object mask in practice.
Similar to the ego-motion estimation, an initial object motion model is generated for initialisation.
The model with most inliers is refined using~\eqref{eq:cost_3} to get the final object motion and the best point matching.

\begin{figure*}[h]
    \centering
    \subfigure{\includegraphics[height=1.3in]{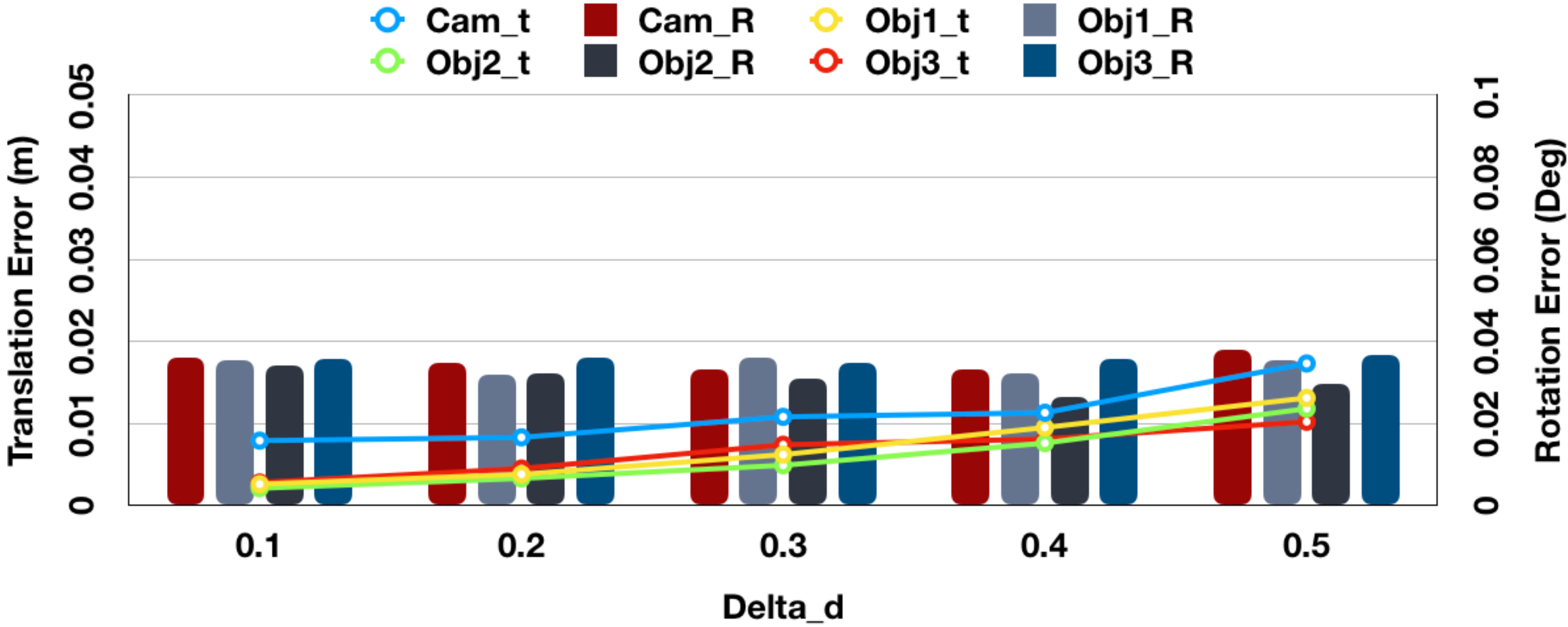}}
    \hspace{2em}
    \subfigure{\includegraphics[height=1.3in]{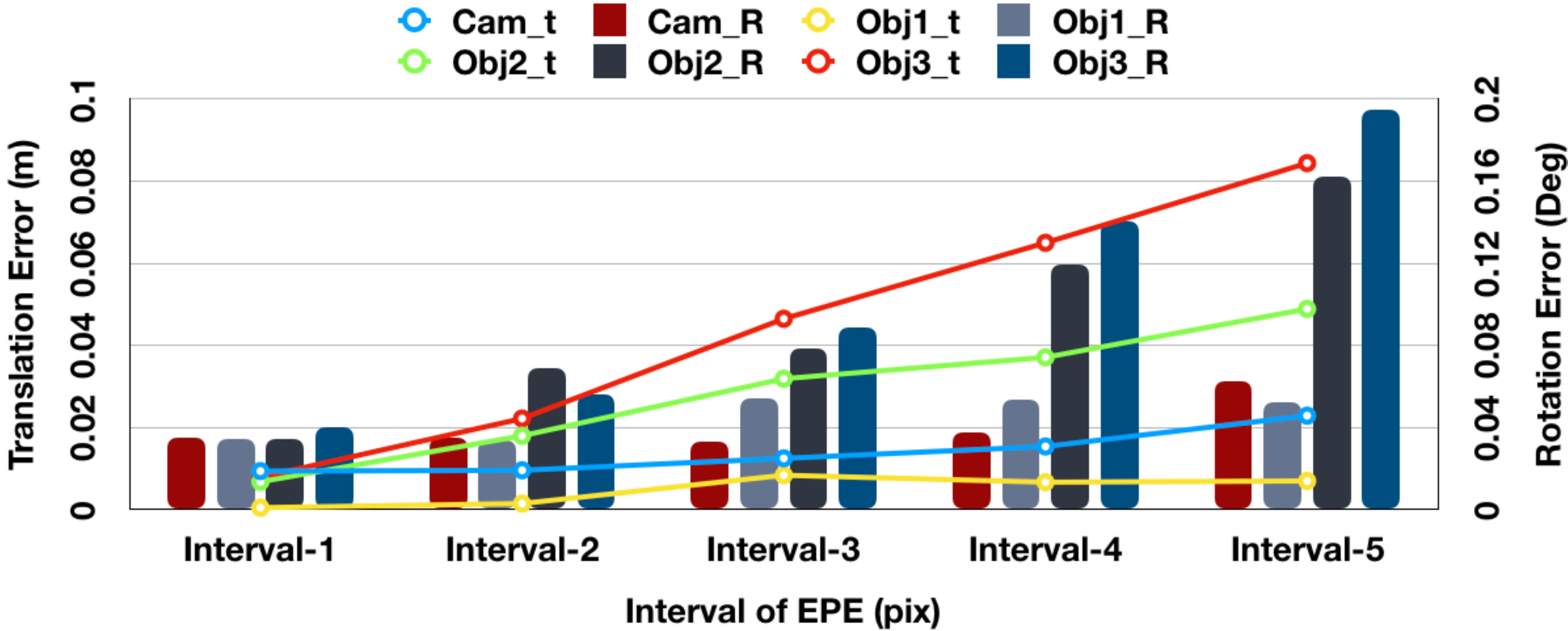}}
    \caption{Average error of rigid motion with regard to noise level of depth (left), and to End-point Error of optical flow (right). Curves represent translation error that are corresponding to left-Y axis, and bars represent rotation error that are corresponding to right-Y axis.}
    \label{fig:mot_analy}
    \vspace{-.5cm}
\end{figure*}

\section{Experiments}
\label{sec:experi}

\noindent In this section, experimental results on two public datasets are demonstrated.
For detailed analysis  we use virtual KITTI dataset~\cite{Gaidon2016CVPR}, which provides ground truth of ego/object poses, depth, optical flow and instance level object segmentation.
KITTI tracking dataset~\cite{Geiger2012CVPR} is used to show the applicability of our algorithm in real life scenarios.
We adopt a learning-based method, Mask R-CNN~\cite{he2017iccv}, to generate object segmentation in both datasets.
The model of this method is trained on COCO dataset~\cite{lin2014eccv}, and it is directly used without fine-tuning.
For dense optical flow, we use a state-of-the-art method, PWC-Net~\cite{sun2018cvpr}.
The model is trained on FlyingChairs dataset~\cite{mayer2016cvpr}, and then fine-tuned on Sintel~\cite{butler2012eccv} and KITTI training datasets~\cite{Geiger2012CVPR}.
Feature detection is done using FAST~\cite{rosten2006eccv}.

We use pose change error to evaluate the estimated \SE motion, i.e., given ground truth motion ${\bf{X}}$ and estimated $\hat{\bm{X}}$, where \mbox{${\bf{X}}\in \SE$} can be either camera or object motion.
The pose change error is obtained as: \mbox{${\bf{E}} = \hat{\bm{X}}^{-1}\:\bm{X}$}.
Translation error ${E}_{t}$ is computed as the $L_2$ norm of translational component in $\bf{E}$.
Rotation error ${E}_{R}$ is measured as the angle in axis-angle representation of rotation part of $\bf{E}$.
We also evaluate object velocity error.
According to~\cite{Chirikjian2017idetc}, given an object motion $\bm{H}$, the object velocity $v$ can be calculated as: \mbox{$v = ||\bm{t} - (\bm{I}-\bm{R})\: \bm{c}||$} where $\bm{R}$ and $\bm{t}$ are the rotation and translation part of the motion of points in global reference frame.
$\bm{I}$ is identity matrix and $\bm{c}$ is centroid of object.
Then error of velocity $E_v$ between estimated $\hat{v}$ and ground truth $v$ can be represented as: \mbox{$E_{v} = {|\hat{v}-v|}$}.
The optical flow is evaluated using end-point error (EPE)~\cite{sun2014ijcv}.

\begin{table}[h]
  \centering
  \fontsize{7}{8}\selectfont
  \caption{Average optical flow end-point error (EPE) of static background and objects in S$18$-F$124$-$134$.}
  \label{tab:flow_err}
 \begin{tabular}{ccccc}
  \toprule
                           &Static  &Obj1   &Obj2    &Obj3    \cr
  \midrule
  Object Distance (m)      &$-$     &7.52   &16.52   &24.67   \cr
  Object Area (\%)         &$-$     &6.29   &0.73    &0.29    \cr
  EPE X-axis (pix)         &1.34    &0.35   &0.34    &0.15    \cr
  EPE Y-axis (pix)         &0.27    &0.24   &0.22    &0.18    \cr
  \bottomrule
 \end{tabular}
 \vspace{-.2cm}
\end{table}

\subsection{Virtual KITTI Dataset}

This dataset is used to analyse the influence of the optical flow and depth accuracy on the estimation of the ego and object motion.
Moving objects appears scatteredly within a sequence, which makes it hard to perform in-depth tests using the whole sequence.
Therefore, we selected a representative set that contains multiple moving objects for analysis.
The set is part of the sequence $18$ and the frame IDs are between $124$-$134$ (S$18$-F$124$-$134$).
It contains $10$ frames of the agent car with camera moving forward, and three observed vehicles.
Two of them are moving alongside in the left lane, with one closer to the camera and the other farther.
The third car is moving upfront and it is furthest from the camera.

\begin{table}[h]
  \centering
  \fontsize{7}{8}\selectfont
  \caption{Average error of object motions of different sets.}
  \label{tab:mot_all}
 \begin{tabular}{cccccc}
  \toprule
  \multicolumn{2}{c}{}   &\multicolumn{2}{c}{Motion only}       &\multicolumn{2}{c}{Joint}       \cr
  \midrule
  \multicolumn{2}{c}{}   &$E_{t}$ (m)  &$E_{R}$ (deg)           &$E_{t}$ (m)    &$E_{R}$ (deg)   \cr
  \midrule
  S$01$-F$225$-$235$     &Ego    &0.0117       &0.0354          &\bm{$0.0043$}  &\bm{$0.0310$}   \cr
  S$01$-F$410$-$418$     &Obj    &0.0647       &0.2811          &\bm{$0.0470$}  &\bm{$0.2286$}   \cr
  \multirow{3}{*}{S$18$-F$124$-$134$}
                         &Ego    &0.0367       &0.1012          &\bm{$0.0052$}  &\bm{$0.0315$}   \cr
                         &Obj1   &0.0169       &0.1016          &\bm{$0.0132$}  &\bm{$0.0804$}   \cr
                         &Obj2   &0.1121       &0.2720          &\bm{$0.1008$}  &\bm{$0.1907$}   \cr
  \bottomrule
 \end{tabular}
 \vspace{-.2cm}
\end{table}

$\mathbf{Depth}$: Ground truth depth is corrupted with zero mean Gaussian noise, with $\sigma$ following standard depth accuracy of a stereo camera system expressed as: \mbox{$\sigma = \frac {z^2}{f\cdot b}\cdot \triangle{d}$} where $z$ is depth, $f$ focal length, $b$ baseline and $\Delta{d}$ the disparity accuracy. We set $b=0.5$ m and control $\Delta{d}$ to get the noise level of depth. Normally, $\Delta{d}$ varies from $0.1$ to $0.2$ for a standard industrial stereo camera. Fig.~\ref{fig:mot_analy} (left) demonstrates the average error of rigid motion over all selected frames. Note that our algorithm is robust to depth noise within reasonable range.
The translation error grows gradually with the depth error for both camera and objects, but stays in low range (\mbox{$E_{t}<0.02$}m). Rotation error fluctuates slightly but still in low range (\mbox{$E_{R}<0.04$}deg). %

\begin{figure}[h]
 \centering
 \includegraphics[width=1.\columnwidth]{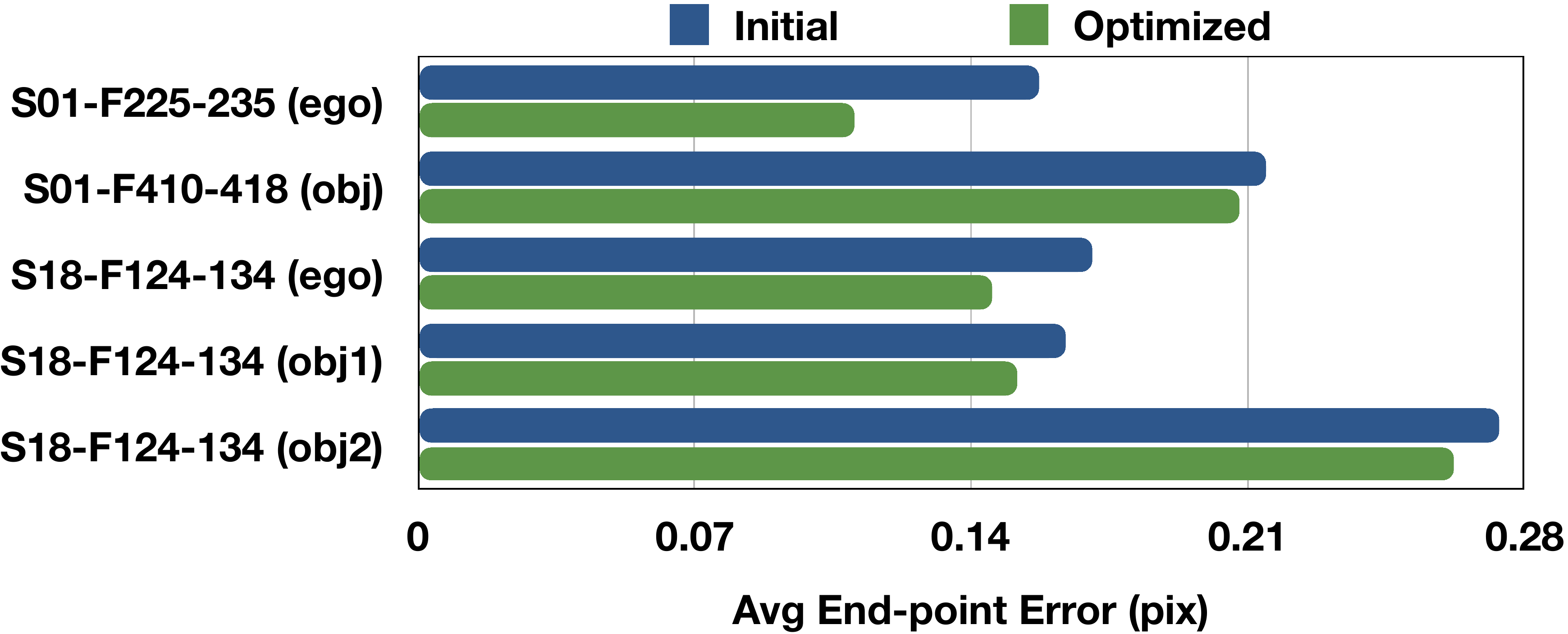}
 \caption{Average end-point error between initial and optimized optical flow, among different tested sets.}
 \label{fig:flow_err}
 \vspace{-.2cm}
\end{figure}

\begin{table*}[!h]
  \centering
  \fontsize{7}{8}\selectfont
  \caption{Average velocity error of sequences with multiple moving objects.}
  \label{tab:velocity}
 \begin{tabular}{cccccccccccc}
  \toprule
  Sequence &\multicolumn{2}{c}{$00$}  &$01$ &$02$ &\multicolumn{2}{c}{$03$} &$04$ &$05$ &$06$ &$18$ &$20$         \cr
  \midrule
  Detected Objects  &van &cyclist &5 cars &6 cars &wagon &suv &20 cars &12 cars &10 cars &18 cars &46 cars  \cr
  \midrule
  Num. of Tracks       &44    &90    &76    &39    &44    &49    &109   &57    &137   &431   &489    \cr
  Avg. Velocity (km/h)    &18.92 &16.06 &14.07 &34.29 &54.44 &52.23 &30.12 &45.22 &32.82 &20.95 &11.73  \cr
  Avg. Error $E_{v}$ (km/h)  &3.04  &2.01  &2.02  &5.22  &2.70  &2.63  &5.13  &5.52  &4.26  &1.96  &2.18  \cr
  \bottomrule
 \end{tabular}
 \vspace{-.2cm}
\end{table*}

\begin{figure*}[!h]
    \centering
    \subfigure{\includegraphics[height=1.4in]{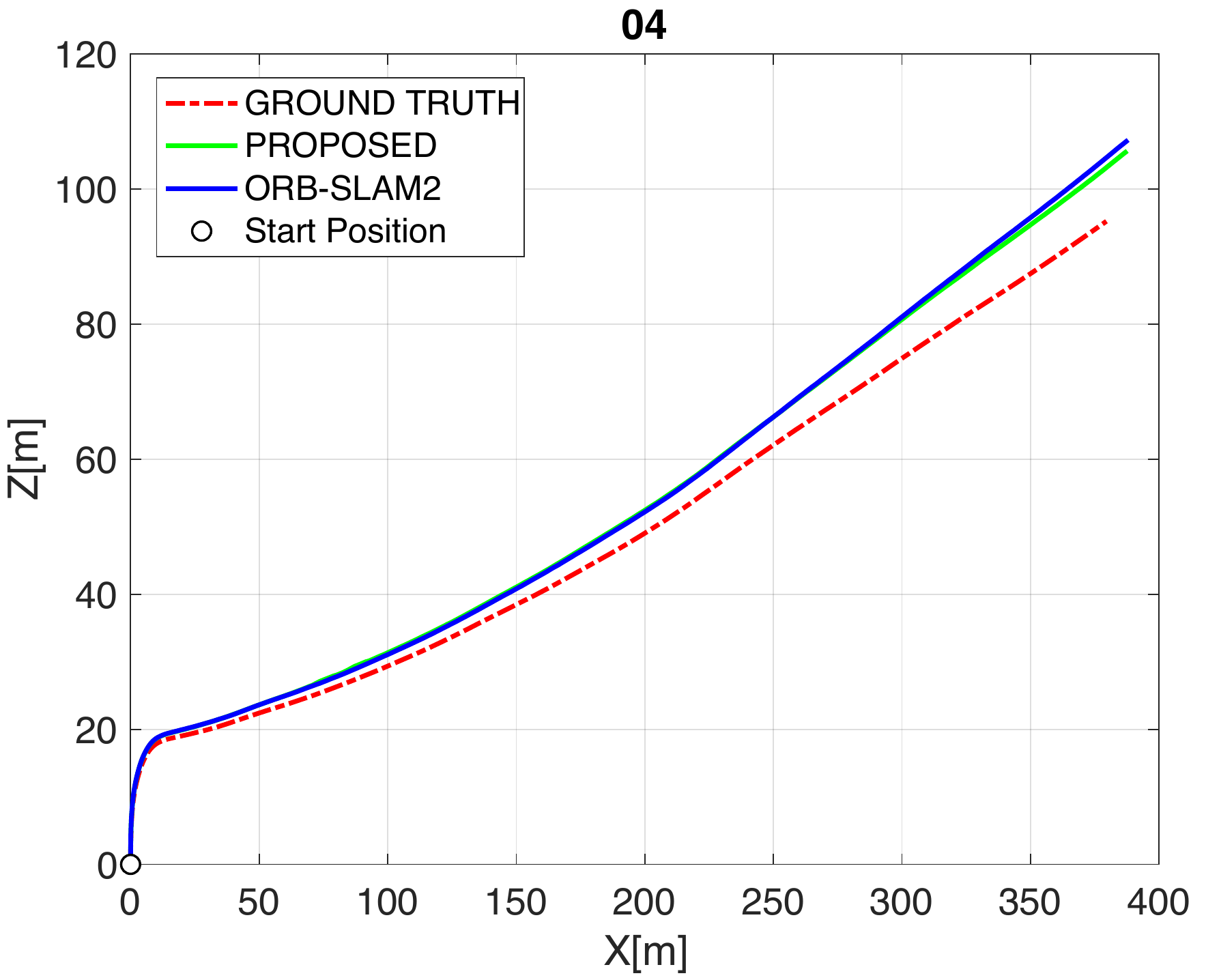}}
    \hspace{4em}
    \subfigure{\includegraphics[height=1.4in]{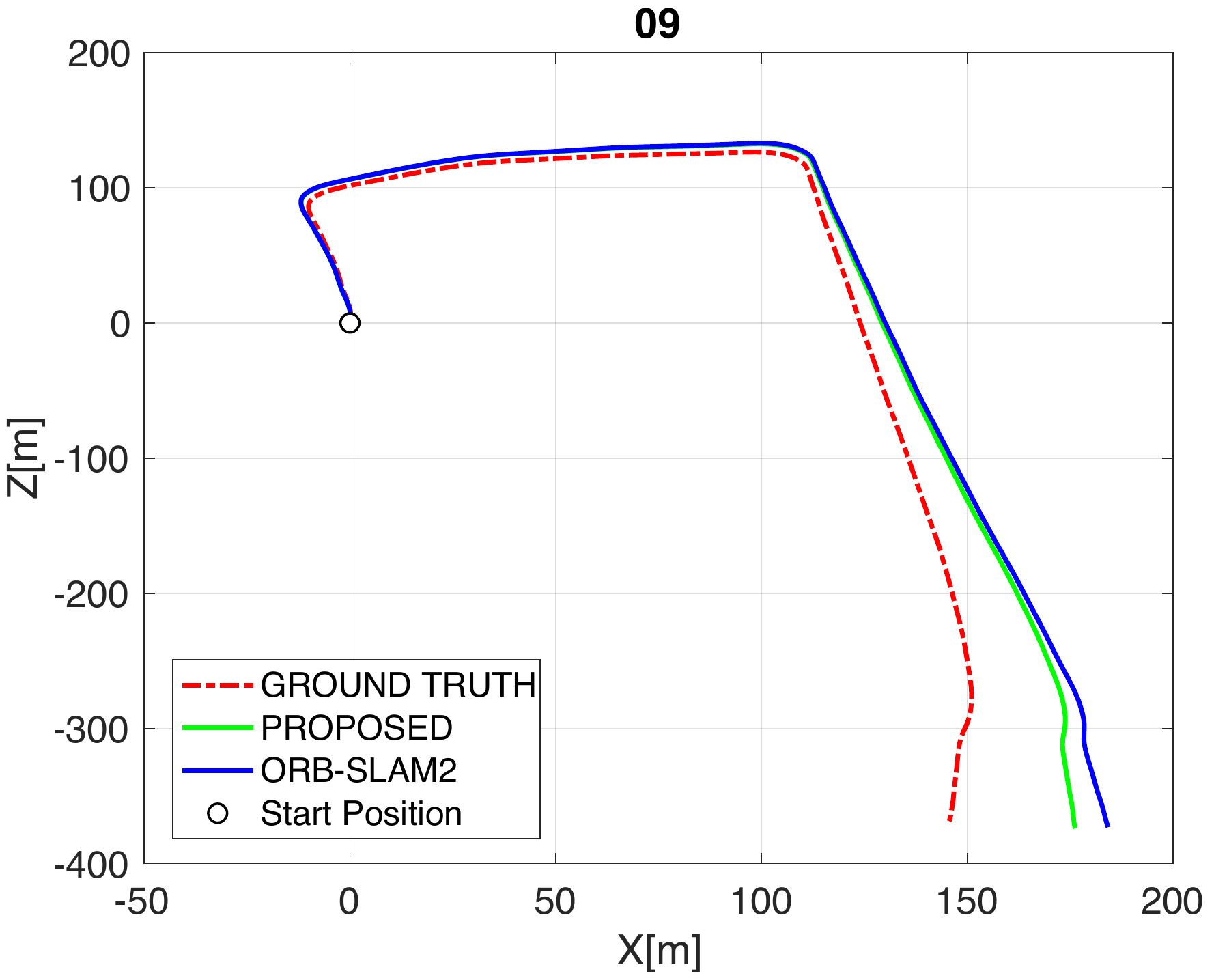}}
    \hspace{4em}
    \subfigure{\includegraphics[height=1.4in]{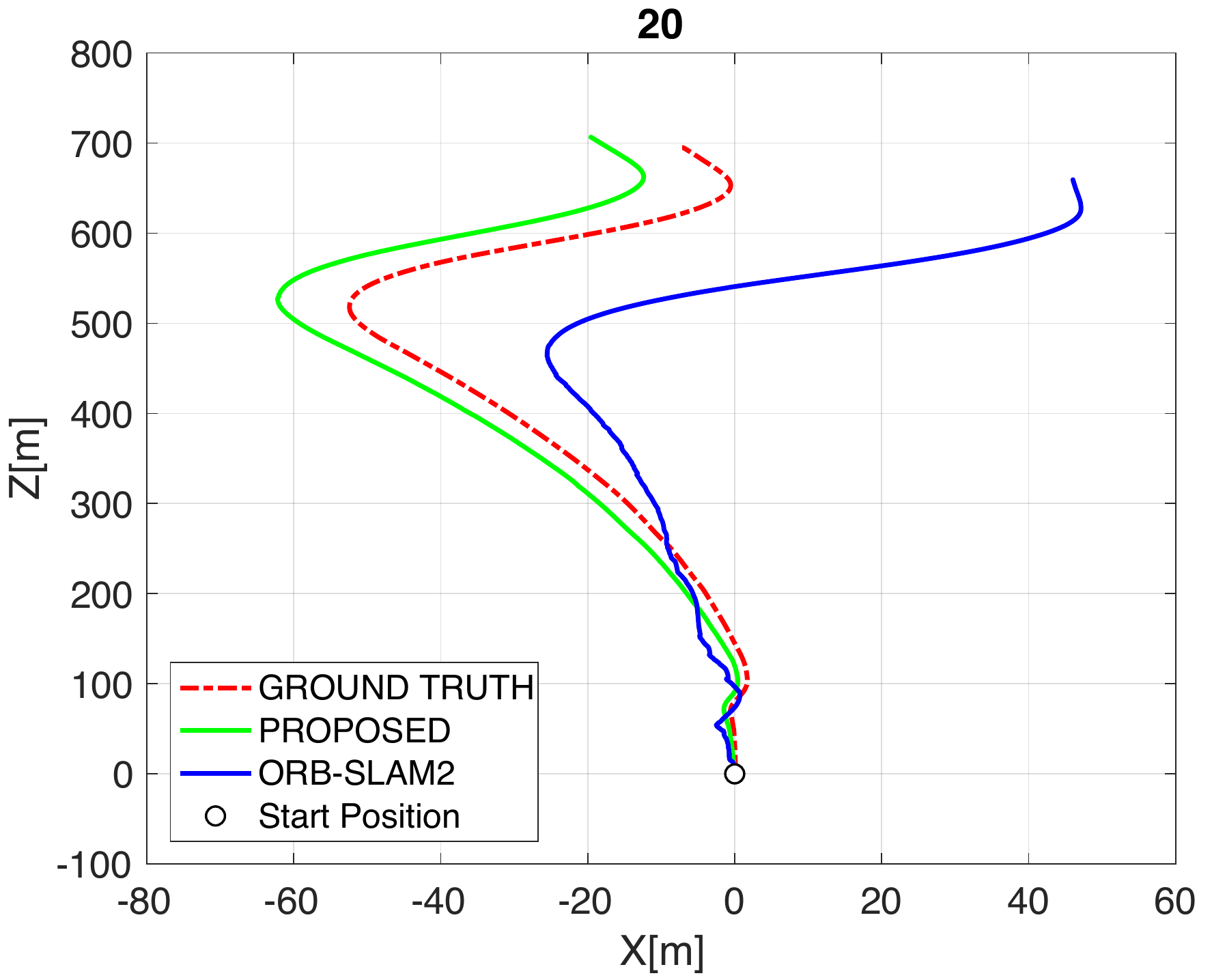}}
    \caption{Top view of camera trajectories of three tested KITTI sequences.}
    \label{fig:cam_trj}
    \vspace{-.5cm}
\end{figure*}

$\mathbf{Optical Flow}$: The ground truth optical flow is corrupted with zero mean Gaussian noise with $\sigma$ decided by the end-point error (EPE). Table~\ref{tab:flow_err} demonstrates average EPE of PWC-Net results for the static and object points among this sequence. Since the errors among static background and objects are different, we set five intervals in increasing order and use these average errors as the middle value. For instance, for static points, $\sigma_{y}$ is set to [\mbox{$0.09$ $0.18$ $0.27$ $0.36$ $0.45$}].

As illustrated in Fig.~\ref{fig:mot_analy} (right) and Table~\ref{tab:flow_err}, camera and object $1$ motion errors are relatively low and stable for different EPEs. However, objects $2$ and $3$ motion errors increase reaching nearly $0.09$ meter in translation and $0.2$ degree in rotation and this is because they are far away from the camera and occupy quite small area of the image (\mbox{$<1\%$}). Consequently, the object motion estimation is sensitive to optical flow error if the objects are not well distributed in the scene. To avoid unreliable estimation, our system addresses only objects within $25$m, and $0.5\%$ image presence.

$\mathbf{Overall}$ $\mathbf{Results}$: Now overall results without ground truth are demonstrated. Because vKITTI does not provide stereo images, we can not generate depth map. Instead, we use ground truth depth map and add noise with \mbox{$\Delta{d}=0.2$}.

As the objects in \mbox{S$18$-F$124$-$134$} are mainly translating, we introduce two more sets with obvious rotation. One of them (\mbox{S$01$-F$225$-$235$}) contains the agent car (camera) turning left into the main street. The other (\mbox{S$01$-F$410$-$418$}) contains static camera observing one car turning left at the crossroads. To prove the effectiveness of jointly optimising motion and optical flow, we set a baseline method that only optimises for motion (Motion Only) using~\eqref{eq:cost_1} for camera or~\eqref{eq:cost_2} for object, and the improved method that optimises for both motion and optical flow with prior constraint (Joint) using~\eqref{eq:cost_3}.

As illustrated in Table~\ref{tab:mot_all}, optimising for the optical flow jointly with the $\SE$ motions improve the results, about $300\%$ for the camera motion, and $10$-$20\%$ on object motion. Besides, the corresponding optical flow error after optimisation is also reduced, see Fig.~\ref{fig:flow_err}.

\subsection{Real KITTI Dataset}

In KITTI tracking dataset, there are $21$ sequences with ground truth camera and object poses. For camera motion, we compute the ego-motion error on all the sequences ($12$ in total) except the ones that the camera is not moving at all. We also generate results of a state-of-the-art method, ORB-SLAM2~\cite{mur2017tor} for comparison. Fig.~\ref{fig:cam_trj} illustrates the camera trajectory results on three sequences. Compared with ORB-SLAM2, our proposed method is able to produce smooth trajectories that are more consistent with the ground truth, given the fact that our method conducts only frame-by-frame tracking, while ORB-SLAM2 integrates more complex modules, such as local map tracking and local bundle adjustment. In particular, the result of Seq.~$20$ in Fig.~\ref{fig:cam_trj} (right) shows that ORB-SLAM2 obtains bad estimates in first half of the sequence, mainly because this part contains dynamic scenes of traffic on the highway. Nevertheless, our algorithm is robust against this case. Table~\ref{tab:cam_err} illustrates average motion error over all the $12$ tested sequences. The results prove our improved performance over ORB-SLAM2.

\begin{table}[h]
  \centering
  \fontsize{7}{7}\selectfont
  \caption{Average ego-motion error over $12$ tested sequences.}
  \label{tab:cam_err}
 \begin{tabular}{ccc}
  \toprule
                     &PROPOSED          &ORB-SLAM2      \cr
  \midrule
  $E_{t}$ (m)        &\bm{$0.0642$}     &0.0730         \cr
  $E_{R}$ (deg)      &\bm{$0.0573$}     &0.0622         \cr
  \bottomrule
 \end{tabular}
 \vspace{-.4cm}
\end{table}

For object motion, we demonstrate the results of object velocity error among $9$ sequences that contains considerable number of moving objects, since vehicle velocity is important information for autonomous and safety driving applications. As demonstrated in Table~\ref{tab:velocity}, the number of tracks refers to how many frames those objects are being tracked. This indicates our pipeline is able to simultaneously and robustly track multiple moving objects for long distances. The average velocity error $E_{v}$ is computed over all the tracks among one or all objects (see the second row in Table~\ref{tab:velocity}). Overall, our method gets around $2$-$5$km/h error, which is considerably accurate for the velocity ranging from $11$-$55$km/h.

The computational cost of our algorithm is around $6$fps when run on an i$7$ $2.6$Ghz laptop. The main cost lies in denser points tracking on multiple objects. This can be improved by employing parallel implementation to achieve real-time performance.

\section{Conclusion}

In this paper we present a novel framework to simultaneously track camera and multiple object motions. The proposed framework detects moving objects via combining instance-level object segmentation and scene flow, and tracks them over frames using optical flow. The \SE motions of the objects, as well as the camera are optimised jointly with the optical flow in a unified formulation. We carefully analyse and test our approach on virtual KITTI dataset, and demonstrate its effectiveness. Furthermore, we perform extensively test on the real KITTI dataset. The results show that our method is able to obtain robust and accurate camera trajectories in dynamic scene, and track the velocity of objects with high accuracy. Further work will integrate the proposed motion estimation within a SLAM framework.



\section*{ACKNOWLEDGMENT}

\noindent This research is supported by the Australian Research Council through the Australian Centre of Excellence for Robotic Vision (CE140100016).

\balance
\bibliographystyle{IEEEtran}
\bibliography{refs/references}

\end{document}